\documentclass[10pt,twocolumn,letterpaper]{article}

\usepackage[pagenumbers]{style}
\usepackage[dvipsnames]{xcolor}

\definecolor{cvprblue}{rgb}{0.21,0.49,0.74}
\usepackage[pagebackref,breaklinks,colorlinks,citecolor=cvprblue]{hyperref}
\usepackage{multirow}
\usepackage{colortbl}
\usepackage{url}

\title{Multi-modal In-Context Learning Makes an Ego-evolving Scene Text Recognizer}

\author{ Zhen Zhao$^{1,2,*}$, Jingqun Tang$^{2,\dag}$, Chunhui Lin$^2$, Binghong Wu$^2$,  Can Huang$^{2}$, \\Hao Liu$^2$, Xin Tan$^1$, Zhizhong Zhang$^1$, Yuan Xie$^{1,\dag}$\\
$^1$ East China Normal University \ $^2$ByteDance \\
{\tt\small \{51255901056\}@stu.ecnu.edu.cn, \{zzzhang,xtan,yxie\}@cs.ecnu.edu.cn} \\
{\tt\small \{tangjingqun,linchunhui.26,wubinghong,haoliu.0128,can.huang\}@bytedance.com}
}

\begin{document}
\maketitle
\begin{abstract}

Scene text recognition (STR) in the wild frequently encounters challenges when coping with domain variations, font diversity, shape deformations, {\it etc.} A straightforward solution is performing model fine-tuning tailored to a specific scenario, but it is computationally intensive and requires multiple model copies for various scenarios. Recent studies indicate that large language models (LLMs) can learn from a few demonstration examples in a training-free manner, termed ``In-Context Learning" (ICL). 
Nevertheless, applying LLMs as a text recognizer is unacceptably resource-consuming. Moreover, our pilot experiments on LLMs show that ICL fails in STR, mainly attributed to the insufficient incorporation of contextual information from diverse samples in the training stage. To this end, we introduce E$^2$STR, a STR model trained with context-rich scene text sequences, where the sequences are generated via our proposed in-context training strategy. E$^2$STR demonstrates that a regular-sized model is sufficient to achieve effective ICL capabilities in STR. Extensive experiments show that E$^2$STR exhibits remarkable training-free adaptation in various scenarios and outperforms even the fine-tuned state-of-the-art approaches on public benchmarks. The code is released at {\color{red} https://github.com/bytedance/E2STR}.

\end{abstract}
\renewcommand{\thefootnote}{}
\footnotetext{$^{\dag}$ Corresponding authors.}
\footnotetext{$^{*}$This work is done when Zhen Zhao is an intern at ByteDance.}
\section{Introduction}
\label{intro}

Scene Text Recognition (STR) is a fundamental task in computer vision, with extensive applications in several domains such as autonomous driving \cite{driving-1},  augmented reality \cite{augmented-1,augmented-2}, industrial print recognition \cite{industrial-print-1} and visual understanding \cite{visual-understanding-1}. 

Current progress in STR \cite{Union14M,parseq,satrn,nrtr} has demonstrated remarkable performance in numerous scenarios. 

\begin{figure}[t]
        \centering
        \includegraphics[width=0.48\textwidth]{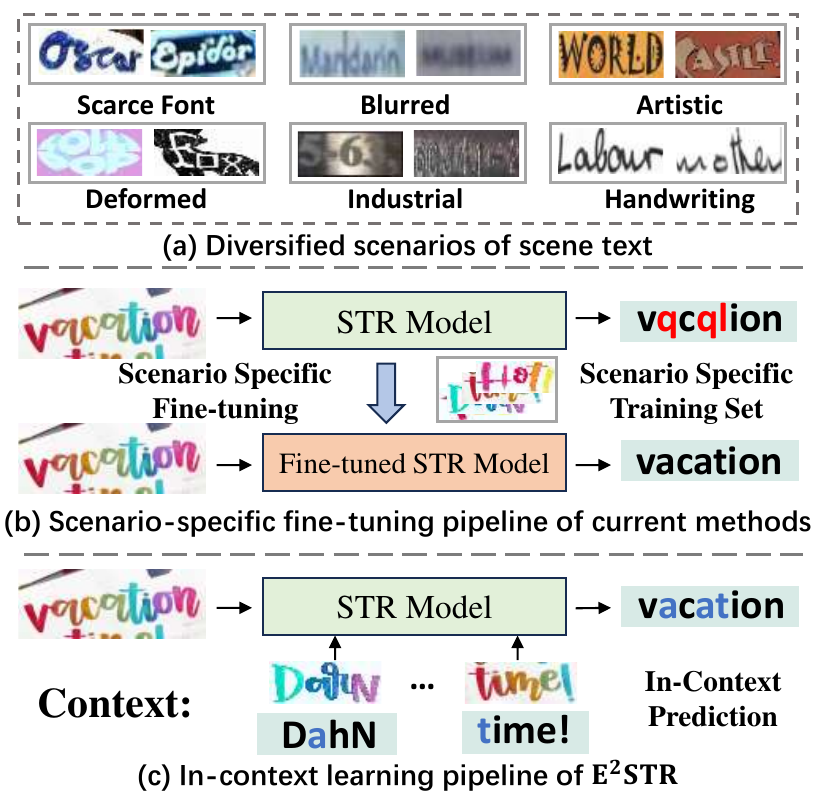}
        \caption{Demonstration of real-world scene text scenarios and the adaptation pipeline. (a) Diversified scenarios of scene text in the real world. (b) The adaptation pipeline of current methods. They typically have to fine-tune upon a trained STR model with the training set, under a specific scenario. (c) The adaptation pipeline of our proposed E$^2$STR. Our method automatically selects in-context prompts and performs training-free adaptation when faced with novel scenarios. {\color{Blue} Blue} characters denote ambiguous scene text that is easily misrecognized.}
        \label{fig: motivation}
\end{figure}

However, as shown in Figure \ref{fig: motivation} (a), STR models are supposed to perform robustly over diversified scenarios in the real world, where the scene text is hard to recognize because of domain variation, font diversity, shape deformation, etc. As shown in Figure \ref{fig: motivation} (b), a straightforward solution involves collecting the corresponding data and then fine-tuning the model for the specific scenario \cite{Union14M,parseq,satrn}. This process is computationally intensive and requires multiple model copies for diverse scenarios. 

The development of a comprehensive and reliable STR model that can effectively handle many real-world scenarios remains a significant challenge.

Fortunately, plenty of studies \cite{gpt3,flamingo,frozen,otter} have shown that Large Language Models (LLMs) can easily adapt without additional training. This adaptation is achieved by leveraging only a handful of input-label pairs as context (prompting information), a phenomenon known as ``In-Context Learning" (ICL). The advantages of ICL inspire our interest in implementing it in STR, such that by fetching a few in-context prompts, a single model can be rapidly adapted to various scenarios without fine-tuning.

However, the equipment of ICL in STR still poses challenges under the existing circumstances.
Firstly, it is deemed excessively costly to apply Multi-Modal Large Language Models (M-LLMs) with billions of parameters as a scene text recognizer. And the ICL capabilities in regular-sized models have been barely explored currently.

\begin{figure}[t]
        \centering
        \includegraphics[width=0.48\textwidth]{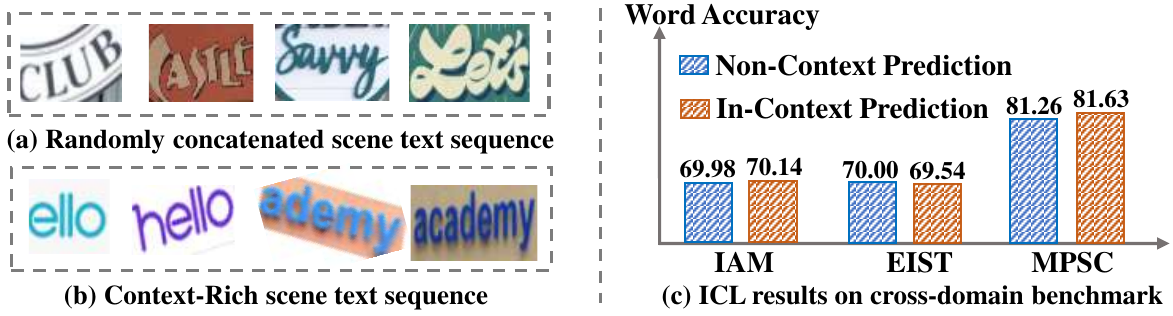}
        \caption{Our pilot experiments. (a) The randomly concatenated scene text sequence. (b) Our proposed context-rich scene text sequence. (c) By training an STR model based on the randomly concatenated scene text sequence, we evaluate the model on three cross-domain datasets.}

        \label{fig: pilot experiment}
\end{figure}

Secondly, it is hard to acquire ICL capabilities for a STR model with current training strategies. Previous studies have observed that sending image-text sequences for training would naturally endow ICL for M-LLMs \cite{frozen,flamingo,otter}, while such a phenomenon is hard to achieve in STR. As shown in Figure \ref{fig: pilot experiment} (a), we generate sequential training data by randomly concatenating scene text samples. {This practice fails as the trained model does not exhibit any performance improvement even when provided with in-domain prompts (Figure \ref{fig: pilot experiment} (c)).}
The major cause of this failure is the lack of \emph{context} in the generated scene text sequences during the training phase. The arbitrary concatenation of scene text does not provide any contextual information ({\it i.e.}, sample connections) between different samples (Figure \ref{fig: pilot experiment} (a)). Consequently, the model lacks the ability to effectively use information derived from in-context prompts(Figure \ref{fig: pilot experiment} (c)), which implies that \emph{in-context training} is essentially important for the effective implementation of ICL in STR.

Based on the above analysis, we propose E$^2$STR (Ego-Evolving STR), 
a paradigm that facilitates adaptation across diverse scenarios in a training-free manner.
Specifically, we propose an in-context training strategy, which enables the model to {exploit contextual information} from the generated context-rich scene text sequences (Figure \ref{fig: pilot experiment} (b)). The context-rich scene text sequences are formed using our ST-strategy, which involves random \textbf{S}plitting and \textbf{T}ransformation of scene text, hence generating a set of ``sub-samples". The sub-samples are inner-connected in terms of both visual and linguistic aspects. 
In the inference stage, E$^2$STR fetches in-context prompts based on visual similarities, {and utilizes the prompts to assist the recognition}, shown in Figure \ref{fig: motivation} (c). In practice, it is found that with proper training and inference strategies, ICL capabilities can also be observed in regular-sized STR models (hundreds of millions of parameters).

Finally, the proposed E$^2$STR effectively captures contextual information from the in-context prompts and performs rapid adaptation in various novel scenarios in a training-free manner (Please refer to Section \ref{sec: Main Results}). On common benchmarks, E$^2$STR achieves SOTA results, with an average improvement of $0.8$$\%$ over previous methods and 1.1$\%$ over itself without ICL. Most importantly, when evaluated on unseen domains, E$^2$STR achieves impressive performance with only a few prompts, even outperforming the fine-tuning results of SOTA methods by $1.2$$\%$.  
Our contributions are summarized below:

(1) We propose E$^2$STR, a robust STR paradigm that can perform rapid adaptation over diverse scenarios in a training-free manner.

(2) We provide an in-context training strategy for equipping STR models with ICL capabilities, as well as an in-context inference strategy for STR models to leverage contextual information from in-context prompts.

(3) We demonstrate that ICL capabilities can be effectively incorporated into regular-sized STR models via appropriate training and inference strategies. 

(4) Extensive experiments show that E$^2$STR exceeds state-of-the-art performance across diverse benchmarks, even surpassing the fine-tuned approaches in unseen domains.

\section{Related Work}

\subsection{Scene Text Recognition}

Recent years have witnessed extensive studies in STR, which can be generally divided into Language-free methods and Language-aware methods.

\noindent \textbf{Language-free STR.} Language-free models directly utilize visual features for prediction, without considering the relationship between the characters. In this branch CTC-based \cite{ctc} methods \cite{ctc1,ctc2} play the most prominent part. They typically consist of a CNN for feature extraction and an RNN for sequential feature processing, which are trained end-to-end with the CTC loss \cite{ctc}. Other methods like \cite{segmentation1,segmantation2} focus on treating STR as a character-level segmentation task. The lack of linguistic information limits the application of language-free methods in scenarios with occluded or incomplete characters.

\noindent \textbf{Language-aware STR.} Language-aware models leverage linguistic information to assist the recognition, typically utilizing an external language model (LM) \cite{SRN,abinet} or training internal LMs \cite{internal1,internal2,aster}. SRN \cite{SRN} and ABINet \cite{abinet} feed visual predictions to an external LM for linguistic refinement. The direct application of an external LM without considering visual features leads to possible erroneous correction. On the other hand, methods like PARSeq \cite{parseq} and MAERec \cite{Union14M} implicitly train an internal LM in an auto-regressive manner, which have achieved decent performance. In this paper we base our model on the language-aware design, training a transformer-based language decoder inner-connected with the vision encoder.

\subsection{Multi-Modal In-Context Learning}

Recent large language models (LLMs) \cite{gpt3,OPT} have demonstrated their excellent few-shot adaptation capabilities. By concatenating a few examples with the input as the prompt at reference time, LLMs quickly adapt to novel tasks without parameter updating. This phenomenon introduces a novel learning paradigm termed ``In-Context Learning". Meanwhile, unlike LLMs, vision-language models (VLMs) struggle to understand complex multi-modal prompts \cite{mmicl1}. A large set of approaches \cite{mmicl2,mmicl3,mmicl4,icl-d3ie} have been proposed to empower VLMs with multi-modal in-context learning (M-ICL) capabilities, but they typically utilize vision models (like image caption models) to translate images to text \cite{mmicl3,mmicl4,mmicl7}, or view the LLM as a scheduler learning to call vision experts based on a few examples \cite{mmicl2}. These approaches do not truly establish a VLM with M-ICL capabilities. Recently, several work \cite{frozen,flamingo,otter} proposes to train VLMs with sequential multi-modal data, and have achieved great success in prompting VLMs with multi-modal examples. In this paper, we aim to train a scene text recognizer equipped with M-ICL capabilities based on this sequential training paradigm. We demonstrate that the arbitrary concatenation of scene text fails as stated above, which motivates us to generate context-rich scene text sequences.
\begin{figure*}[t!]
        \centering
        \includegraphics[width=0.98\textwidth]{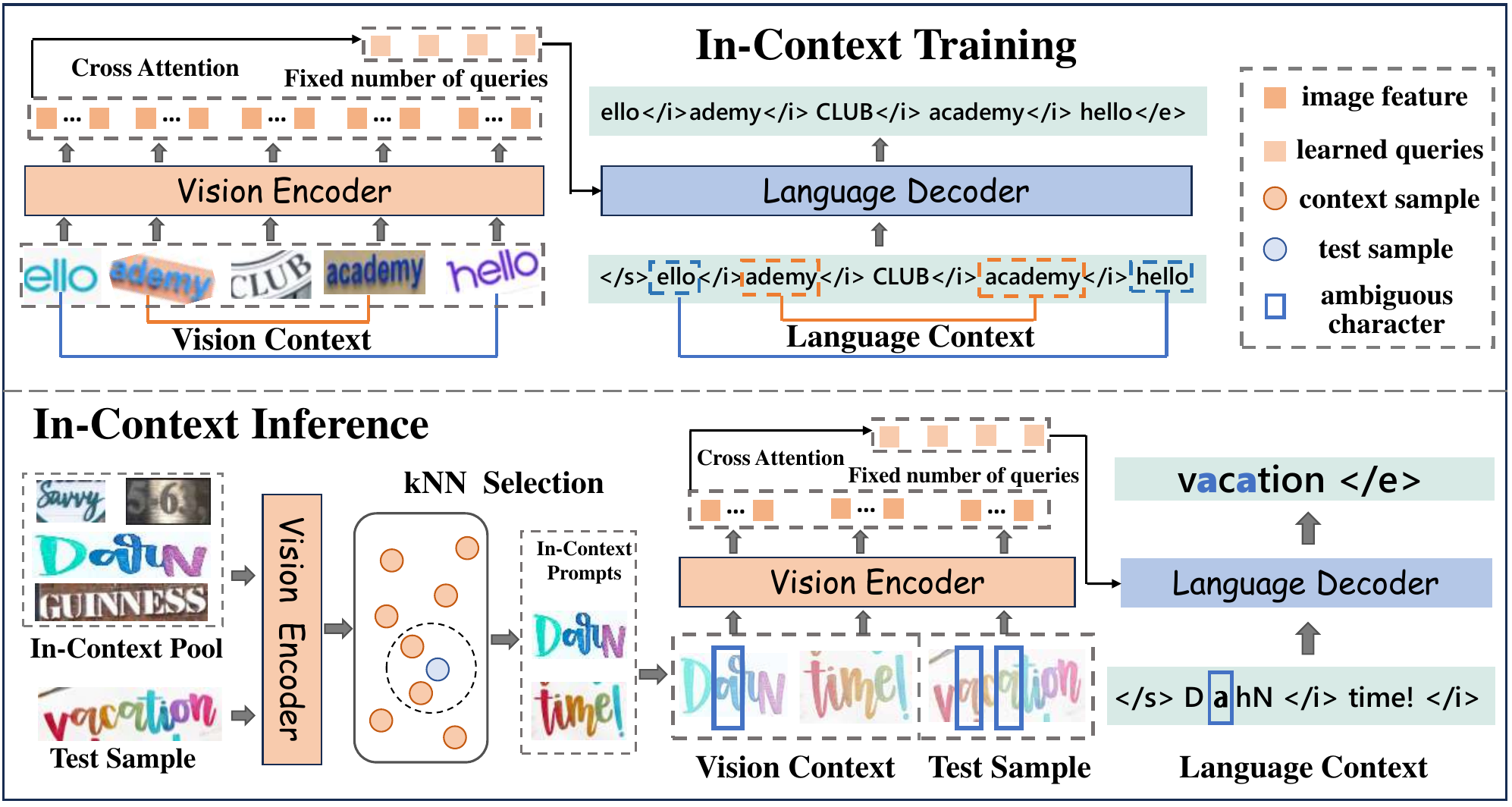}
        \caption{ Pipeline of our E$^2$STR. Top: E$^2$STR is trained with our in-context training strategy to obtain the ICL capability. Down: During inference, E$^2$STR selects in-context prompts based on a kNN strategy, then the test sample grasps context information from the prompts to assist the recognition. Specifically, the ambiguous character ``a" in the test sample is easily misrecognized as ``q". With the vision-language context produced by the in-context prompts ({\it i.e.}, ``a" in the first in-context prompt), E$^2$STR rectifies the result. Note that in practice the in-context pool maintains image tokens and thus does not need to go through the vision encoder.}
        \label{algorithm}
\end{figure*}

\section{Methodology}

\subsection{Preliminary of Multi-Modal In-Context Learning}

Multi-modal in-context Learning enables M-LLMs to perform quick adaptation in downstream tasks in a training-free manner, hence eliminating the redundant computation and time expenses of fine-tuning. In this subsection, we introduce how to formulate multi-modal in-context learning for addressing the STR task.

For a scene text tuple ($\textbf{x}, \textbf{y}$) where $\boldsymbol{x}$ is the scene image and $\boldsymbol{y}$ is the ground-truth text, the STR task involves generating the label $\boldsymbol{y}$ by maximizing the conditional probability under the classic auto-regressive paradigm as follows: $p(\boldsymbol{y}|\boldsymbol{x}) = \prod_{l=1}^{L} p(\boldsymbol{y}_l|\boldsymbol{x}, \boldsymbol{y}_{<l})$, where $\boldsymbol{y}_l$ is the $l$-th character in $\boldsymbol{y}$, $\boldsymbol{y}_{<l}$ is the set of preceding characters, and $L$ is the number of characters in $\boldsymbol{y}$.

While previous state-of-the-art studies typically need to fine-tune pre-trained models when confronted with novel scenarios \cite{Union14M,satrn,parseq}, we propose in this study to leverage multi-modal in-context learning to enable STR models to be rapidly adapted across diverse scenarios without fine-tuning. Specifically, we define the probability of generating the target label $\boldsymbol{y}$ for a given image $\boldsymbol{x}$ and the multi-modal context $C$ as follows:
\begin{equation}
    p(\boldsymbol{y}|\boldsymbol{x}, C) = \prod_{l=1}^{L} p(\boldsymbol{y}_l|\{\underbrace{\boldsymbol{x}_1^c,\cdots,\boldsymbol{x}_n^c}_{\text{vision\ context}}; \boldsymbol{x}\},\{\underbrace{\boldsymbol{y}_1^c,\cdots,\boldsymbol{y}_n^c}_{\text{language\ context}}; \boldsymbol{y}_{<l}\}),
\end{equation}
where the context $C=\{(\boldsymbol{x}_1^c,\boldsymbol{y}_1^c), \cdots, (\boldsymbol{x}_n^c,\boldsymbol{y}_n^c)\}$ {is the set of the in-context prompts}, $(\boldsymbol{x}_i^c,\boldsymbol{y}_i^c)$ are the scene image and the ground-truth text of the context prompts, and $n$ is the number of context prompts.

\subsection{Framework Overview and Model Architecture}

Our proposed E$^2$STR consists of three stages. Firstly, E$^2$STR is trained in the standard auto-regressive framework to learn the fundamental STR ability. 

Secondly, as shown in the top of Figure \ref{algorithm}, E$^2$STR is further trained based on our proposed In-Context Training paradigm. In this stage E$^2$STR learns to understand the connection between different samples, allowing it to profit from in-context prompts. 
Finally, as shown in the bottom of Figure \ref{algorithm}, E$^2$STR fetches in-context prompts based on visual similarity during inference, allowing the test sample to absorb context information.

As shown in the top of Figure \ref{algorithm}, the model architecture of E$^2$STR consists of a vision encoder and a language decoder. The vision encoder receives image inputs and the language decoder processes text inputs in an auto-regressive manner. Following \cite{flamingo}, a set of cross attention layers are utilized to bridge the output tokens of the vision encoder and the language decoder. Under the ICL framework, the vision encoder receives numerous images as input. To control the length of the vision token sequence, a fixed number of query tokens are learned by performing cross attention against the output tokens of the vision encoder.

\begin{figure}[t]
        \centering
        \includegraphics[width=0.48\textwidth]{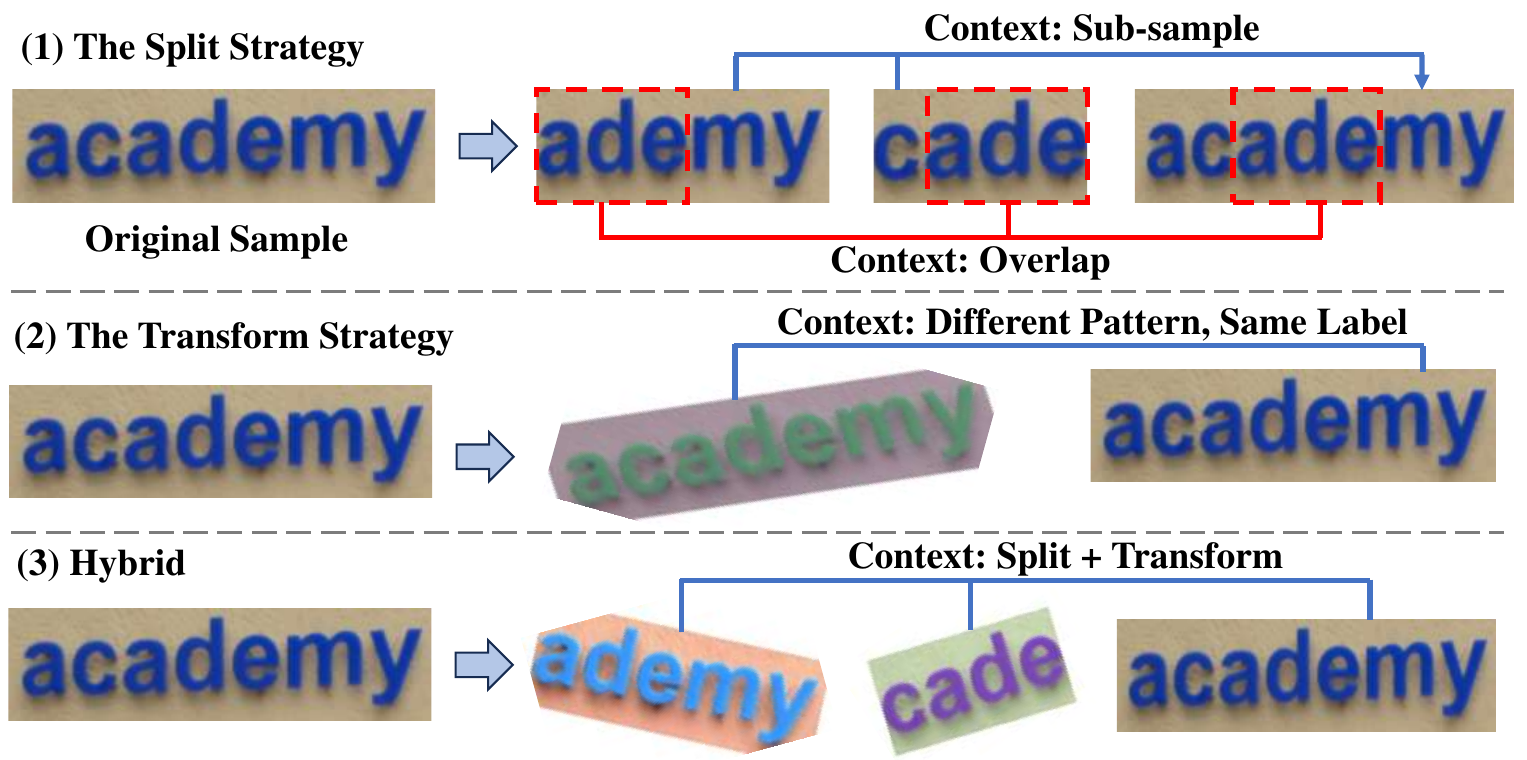}
        \caption{Illustration of the split strategy, the transform strategy, and how we hybrid them in practice.}
        \label{strategy}
\end{figure}

\subsection{Training Strategy}

Our training process is split into two stages: vanilla STR training and in-context STR training.

\subsubsection{Vanilla Scene Text Recognition Training}

The first training phase seeks to provide E$^2$STR with the fundamental skills in STR. For a scene text tuple $(\boldsymbol{x}, \boldsymbol{y})$ the input to the vision encoder is $\boldsymbol{x}$ and the initial input to the language decoder is a start token $<$/s$>$. The training in this phase makes use of the next-token prediction loss:
\begin{equation}
    \mathcal{L} = \mathbb{E}_{(\boldsymbol{x}, \boldsymbol{y})\sim\mathcal{D}}\left[-\sum_{l=1}^{L}\log p({\boldsymbol{y}}_l | {\boldsymbol{y}}_{<l}, \boldsymbol{x})\right],
\end{equation}
where $\mathcal{D}$ is the training set.

\subsubsection{In-Context Training}

The objective of the in-context training phase is to equip E$^2$STR with the capability of In-Context Learning. As depicted in the top of Figure \ref{algorithm}, the model is trained with context-rich scene text sequences as stated before. In these sequences, we interleave a placeholder $<$/i$>$ in the text for each image. This serves to make the language decoder distinguish between different samples following \cite{flamingo}. In this stage, we propose two strategies to generate context-rich scene text sequences: the Split Strategy and the Transform Strategy (the ST strategy).

\noindent \textbf{The Split Strategy.} As shown in Figure \ref{strategy} (a), when presented with a training tuple $(\boldsymbol{x}, \boldsymbol{y})$, we split the sample and hence generating a set of ``sub-samples". It is evident that the sub-samples exhibit a strong connection to the original training sample. Furthermore, the sub-samples themselves demonstrate interconnectivity as they overlap with one another. Next, we proceed to concatenate the sub-samples with $(\boldsymbol{x}, \boldsymbol{y})$ and additional randomly selected samples to form a context-rich sample sequence.

We randomly shuffle the whole sequence before generating the actual input text ({\it i.e.}, interleaving the $<$/i$>$ token to the text sequence).

In practice, to accurately split the training samples, we synthesize 600k scene text images based on \cite{synthesize} and record the accurate bounding boxes of every single character. Our subsequent experiments show that the synthesized data does not change E$^2$STR's non-context text recognition ability, but the Split Strategy based on them equips E$^2$STR with a strong capability of in-context learning.

\noindent \textbf{The Transform Strategy.} As shown in Figure \ref{strategy} (b), given a training tuple $(\boldsymbol{x}, \boldsymbol{y})$ (whether with character-wise bounding boxes or not), we perform data augmentation (a set of image transformations, {\it e.g.,} color/direction transformations) on $\boldsymbol{x}$. In this way, we also generate a set of sub-samples with the same label but different image patterns from the original sample.

In practice, as depicted in Figure \ref{strategy} (c), we hybrid the above strategies. The training set is formed by concatenating the synthesized data and the original training data used in the first training phase. For the synthesized data with character-wise bounding boxes, both the Split Strategy and the Transform Strategy are utilized. For the original training data, only the Transform Strategy is implemented.

Finally, after generating the sample sequence $(\boldsymbol{X}, \boldsymbol{Y})$, where $\boldsymbol{X}$ is the image sequence and $\boldsymbol{Y}$ is the text sequence, $\boldsymbol{X}$ is fed into the vision encoder, while $\boldsymbol{Y}$ is processed by the language decoder under the auto-regressive framework. The loss function is formulated as:
\begin{equation}
    \mathcal{L}_{(\boldsymbol{X}, \boldsymbol{Y})} = -\sum_{l=1}^{L}\log p({\boldsymbol{Y}}_l | {\boldsymbol{Y}}_{<l}, \boldsymbol{X}_{\leq l}),
\end{equation}
where $\boldsymbol{X}_{\leq l}$ is the set of image tokens preceding token $\boldsymbol{Y}_l$ in the input sequence. 

\subsection{In-Context Inference}
\label{method: Inference}

The In-Context Learning ability is acquired by our E$^2$STR model through the above two-stage training approach. AS shown in the bottom of Figure \ref{algorithm}, when presented with a test image $\boldsymbol{x}$, the framework selects $N$ samples $\{(\boldsymbol{x}_i^c, \boldsymbol{y}_i^c)\}_{i=1}^{N}$ from a in-context pool $\mathcal{D}^c$. The selected samples have the highest visual similarities to $\boldsymbol{x}$ in the latent space. Specifically, we calculate the image embedding $\boldsymbol{I}$ of $\boldsymbol{x}$ by averaging the visual token sequence $Encoder(\boldsymbol{x})$. The in-context prompts are then formed by choosing $N$ samples from $\mathcal{D}^c$, where the image embeddings of these samples have the top-N highest cosine similarity with $\boldsymbol{I}$, {\it i.e.,}
\begin{equation}
    \mathcal{I} = \underset{i\in 1, 2, \cdots, |\mathcal{D}^c|}{\mathrm{argTopN}}\, \frac{\boldsymbol{I}^{\text{T}}\boldsymbol{I}_i^c}{\|\boldsymbol{I}\|_2\|\boldsymbol{I}_i^c\|_2},
\end{equation}
where $\mathcal{I}$ is the index set of the top-N similar samples in $\mathcal{D}^c$, and $\boldsymbol{I}_i^c$ is the image embedding of the $i$-th sample in $\mathcal{D}^c$. The in-context prompts are then defined as:
\begin{equation}
    \boldsymbol{E} = \{(\boldsymbol{x}_i^c, \boldsymbol{y}_i^c)|i\in \mathcal{I}\}.
\end{equation}
As shown in the bottom of Figure \ref{algorithm}, $\boldsymbol{E}$ is concatenated with the test sample $\boldsymbol{x}$ and our in-context prediction is formulated as $p(\boldsymbol{y}|\boldsymbol{E}, \boldsymbol{x})$. 
In practice, 
the in-context pool $\mathcal{D}^c$ retains solely the output tokens generated by the vision encoder, resulting in a highly efficient selection process. Furthermore, because the in-context pool is tiny and we do straight inference without training, the extra consumption is kept to a minimum (Please refer to Section \ref{sec:ablation studies}).
\section{Experiment}

\begin{table*}[t!]
\resizebox{\textwidth}{24mm}{
\begin{tabular}{ccccc|cccccc|cc|c|c}
\hline
                         &                         & \multicolumn{3}{c|}{Regular}                                                               & \multicolumn{6}{c|}{Irregular}                                                                                                                                                          & \multicolumn{2}{c|}{Occluded}                               & Others                       &                              \\ \cline{3-14}
                         &                         & IIIT                         & SVT                          & IC13                         & IC15                         & SVTP                         & CT80                         & COCO                         & CTW                          & TT                           & HOST                         & WOST                         & WordArt                      &                              \\
\multirow{-3}{*}{Method} & \multirow{-3}{*}{Venue} & 3000                         & 647                          & 1015                         & 2077                         & 645                          & 288                          & 9896                         & 1572                         & 2201                         & 2416                         & 2416                         & 1511                         & \multirow{-3}{*}{AVG}        \\ \hline
ASTER \cite{aster}                    & PAMI'18                 & 95.03                        & 89.49                        & 93.79                        & 85.48                        & 82.02                        & 90.28                        & 62.25                        & 76.53                        & 78.69                        & 43.34                        & 64.65                        & 65.59                        & 77.26                        \\
NRTR \cite{nrtr}                     & ICDAR'19                & 97.43                        & 93.82                        & 96.06                        & 85.15                        & 84.03                        & 91.32                        & 65.94                        & 81.74                        & 81.83                        & 50.83                        & 71.52                        & 64.06                        & 80.31                        \\
SAR \cite{sar}                      & AAAI'19                 & 97.70                        & 94.13                        & 96.35                        & 87.47                        & 87.60                        & 93.06                        & 67.41                        & 83.91                        & 86.23                        & 46.36                        & 70.32                        & 72.40                        & 81.91                        \\
SATRN \cite{satrn}                    & AAAI'20                 & 97.83                        & 95.83                        & 97.44                        & 89.46                        & 90.85                        & 96.18                        & 73.06                        & 84.61                        & 87.91                        & 56.71                        & 75.62                        & 75.71                        & 85.10                        \\
ABINet \cite{abinet}                   & CVPR'21                 & 97.90                        & 95.98                        & 96.16                        & 91.66                        & 90.23                        & 93.75                        & 71.46                        & 83.72                        & 86.01                        & 56.54                        & 75.75                        & 75.25                        & 84.53                        \\
PARSeq* \cite{parseq}                   & ECCV'22                 & {\color[HTML]{3166FF} 99.10} & 97.84                        & 98.13                        & 89.22                        & 96.90                        & 98.61                        & -                            & -                            & -                            & -                            & -                            & -                            & -                            \\
MAERec \cite{Union14M}                   & ICCV'23                 & 98.93                        & 97.99                        & {\color[HTML]{3166FF} 98.62} & {\color[HTML]{3166FF} 93.04} & 94.57                        & {\color[HTML]{3166FF} 98.96} & {\color[HTML]{FE0000} 78.84} & {\color[HTML]{3166FF} 88.87} & {\color[HTML]{3166FF} 93.91} & {\color[HTML]{3166FF} 73.97} & {\color[HTML]{3166FF} 85.72} & {\color[HTML]{3166FF} 82.59} & {\color[HTML]{3166FF} 90.50} \\ \hline
E$^2$STR-base                 &                         & {\color[HTML]{3166FF} 99.10} & {\color[HTML]{3166FF} 98.15} & 98.03                        & 92.99                        & {\color[HTML]{3166FF} 96.43} & {\color[HTML]{3166FF} 98.96} & 77.29                        & 88.36                        & 93.46                        & 73.30                        & 85.51                        & 81.47                        & 90.25                        \\
E$^2$STR-ICL                  &                         & {\color[HTML]{FE0000} 99.23} & {\color[HTML]{FE0000} 98.61} & {\color[HTML]{FE0000} 98.72} & {\color[HTML]{FE0000} 93.82} & {\color[HTML]{FE0000} 96.74} & {\color[HTML]{FE0000} 99.31} & {\color[HTML]{3166FF} 78.38} & {\color[HTML]{FE0000} 88.99} & {\color[HTML]{FE0000} 94.68} & {\color[HTML]{FE0000} 74.75} & {\color[HTML]{FE0000} 86.59} & {\color[HTML]{FE0000} 86.17} & {\color[HTML]{FE0000} 91.33} \\ \hline
\end{tabular}
}
\caption{Results on common benchmarks. All methods are trained on the same dataset except for PARSeq. *: PARSeq is trained on its self-collected real-world dataset and we directly {quote} the results from its original paper. {\color{red} Red} and {\color{blue} blue} values denote the best and the secondary performance. {E$^2$STR-base refers to non-context inference.}}
\label{Table:Results on common benchmarks}
\end{table*}

\begin{table}[]
\begin{tabular}{ccc|c|c}
\hline
                              & \multicolumn{2}{c|}{Industrial}                             & Handwriting                  &                              \\ \cline{2-4}
                              & MPSC                         & EIST                         & IAM                          &                              \\
\multirow{-3}{*}{Method}      & 2941                         & 8000                         & 3000                         & \multirow{-3}{*}{AVG}        \\ \hline
ASTER \cite{aster}                         & 63.48                        & 48.76                        & 52.50                        & 54.91                        \\
NRTR \cite{nrtr}                          & 73.24                        & 61.77                        & 59.53                        & 64.85                        \\
{\color[HTML]{000000} SAR} \cite{sar}    & {\color[HTML]{000000} 73.85} & {\color[HTML]{000000} 58.26} & {\color[HTML]{000000} 56.63} & {\color[HTML]{000000} 62.91} \\
{\color[HTML]{000000} ABINet} \cite{abinet} & {\color[HTML]{000000} 75.35} & {\color[HTML]{000000} 62.85} & {\color[HTML]{000000} 61.57} & {\color[HTML]{000000} 66.59} \\
{\color[HTML]{000000} SATRN} \cite{satrn}  & {\color[HTML]{000000} 76.10} & {\color[HTML]{000000} 65.42} & {\color[HTML]{000000} 59.47} & {\color[HTML]{000000} 67.00} \\
{\color[HTML]{000000} MAERec} \cite{Union14M} & {\color[HTML]{3166FF} 81.81} & {\color[HTML]{3166FF} 70.33} & {\color[HTML]{3166FF} 70.27} & {\color[HTML]{3166FF} 74.14} \\ \hline
E$^2$STR-base                      & 81.26                        & 69.66                        & 69.51                        & 73.48                        \\
E$^2$STR-ICL                       & {\color[HTML]{FE0000} 83.64} & {\color[HTML]{FE0000} 76.77} & {\color[HTML]{FE0000} 74.10} & {\color[HTML]{FE0000} 78.17} \\ \hline
\end{tabular}
\caption{Results on cross domain scenarios. Three datasets under two unseen domains are evaluated. {All approaches are evaluated in a training-free manner.}}
\label{Results on few-shot scenarios}
\end{table}

\subsection{Experimental Setup}

\noindent \textbf{Implementation Details.} Following MAERec \cite{Union14M}, we choose Vision Transformer \cite{ViT} pre-trained under the MAE \cite{MAE} framework as the vision encoder. The default language decoder is set as OPT-125M \cite{OPT}. We use the cosine learning rate scheduler without warm-up and the AdamW optimizer with a weight decay of 0.01. We train our model for 10 epochs with an initial learning rate of 1e-4 during the first training stage, and 5 epochs with an initial learning rate of 5e-6 during the second in-context training stage. The training batch size is 64 for the first stage and 8 for the second stage. During inference for E$^2$STR-ICL, we select two in-context prompts based on the kNN selection strategy.

\noindent \textbf{Datasets and Metrics.} We use the real-world training dataset Union14M-L\cite{Union14M} for the two-stage training. The same training dataset (including the synthesized data) is adopted for all compared methods. E$^2$STR is evaluated under various publicly available benchmarks, including Regular Benchmarks IIIT5k \cite{IIIT5k}, SVT \cite{svt}, IC13 \cite{ic13}, Irregular Benchmarks IC15 \cite{ic15}, SVTP \cite{svtp}, CUTE80 (CT80) \cite{cute80}, COCO Text (COCO) \cite{coco-text}, CTW \cite{ctw}, Total Text (TT) \cite{total-text}, Occluded Benchmarks OST (HOST and WOST) \cite{ost} and artistic benchmark WordArt \cite{wordart}. In cross domain scenarios the evaluated datasets including the metal-surface benchmark MPSC \cite{mpsc} and the handwriting benchmark IAM \cite{iam}, as well as a more difficult real-world industrial text recognition dataset EIST (Enhanced Industrial Scene Text) collected by us. EIST is collected from the real-world industrial scenario, which contains 200 training samples and 8000 test samples. We use Word Accuracy\cite{Union14M} as the evaluation metric for all compared methods.

\subsection{Main Results}
\label{sec: Main Results}

\subsubsection{Results on Common Benchmarks}

Table \ref{Table:Results on common benchmarks} presents the performance of E$^2$STR on common benchmarks. We evaluate E$^2$STR on 12 commonly used STR benchmarks and compare with SOTA methods. E$^2$STR-base refers to non-context prediction without prompts. For E$^2$STR-ICL, a tiny in-context pool is maintained by randomly sampling 1000 images from the training data (less than 0.025$\%$ of the number of training samples). As we can see, E$^2$STR-base achieves 90.25$\%$ average word accuracy over 12 datasets, 0.25$\%$ lower than MAERec \cite{Union14M}. However, by fetching in-context prompts and exploiting in-context information, E$^2$STR-ICL achieves an average word accuracy of 91.33$\%$, which is 1.08$\%$ higher than E$^2$STR-base and 0.83$\%$ higher than MAERec. Please note that this improvement is automatic and training-free.

Specifically, on the six traditional STR benchmarks ({\it i.e.}, IIIT, SVT, IC13, IC15, SVTP, and CT80) which have nearly reached saturation in recent years\cite{Union14M}, E$^2$STR still push the performance limit from 97.02$\%$ to 97.74$\%$, leading to a 24$\%$ error rate decrease. On the 6 larger and harder STR benchmarks ({\it i.e.}, COCO Text, CTW, TT, HOST, and WOST), E$^2$STR-ICL outperforms MAERec by 0.94$\%$.

\begin{figure}[t]
        \centering
        \includegraphics[width=0.48\textwidth]{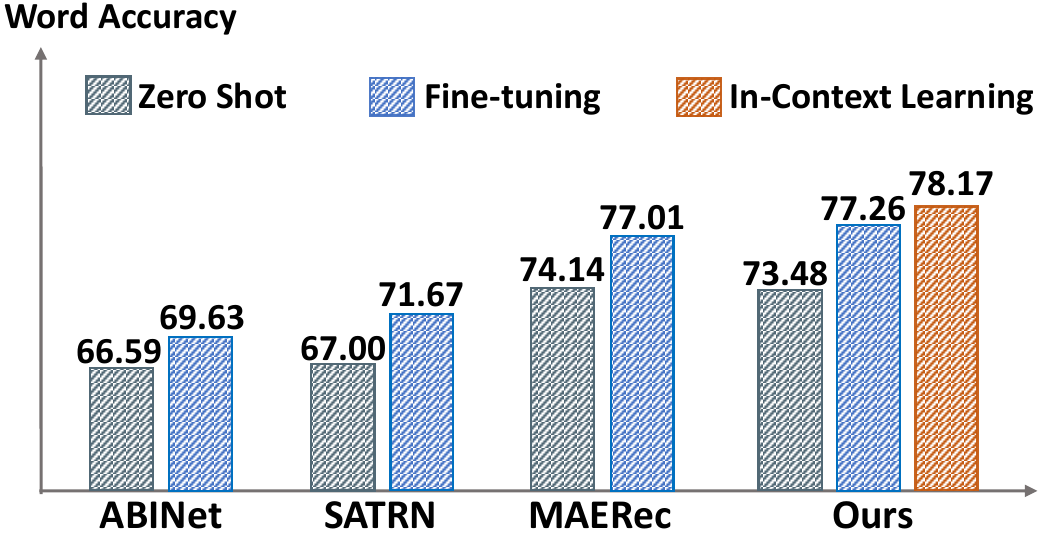}
        \caption{Comparison with the fine-tuned models. We report the average performance on three cross-domain datasets. {Please note that ABINet \cite{abinet}, SATRN \cite{satrn} and MAERec \cite{Union14M} are fine-tuned with the in-domain data, while our E$^2$STR-ICL is training-free.}}
        \label{fig:fine-tuning compare}
\end{figure}

\subsubsection{Results on Cross Domain Scenarios}
\label{sec: Few-shot Scenarios}

We compare with SOTA methods on cross domain benchmarks. Two novel scenarios are introduced: the industrial scenario (MPSC and EIST) and the handwriting scenario (IAM). In each dataset, only 100 training samples are provided. For E$^2$STR-ICL we simply use the training samples as the in-context pool. We compare the training-free results in Table \ref{Results on few-shot scenarios} and the fine-tuning results in Figure \ref{fig:fine-tuning compare}.

As we can see, on both industrial and handwriting scenarios our E$^2$STR-ICL reaches SOTA performance. As shown in Table \ref{Results on few-shot scenarios}, under the training-free constraint E$^2$STR-ICL reaches an average performance of 78.17$\%$, which is 4.69$\%$ higher than E$^2$STR-base and 4.03$\%$ higher than the SOTA method MAERec. Specifically, on EIST and IAM the application of ICL brings a huge improvement of 7.11$\%$ and 4.59$\%$, which demonstrates the extraordinary adaptation ability of E$^2$STR-ICL.

We further compare the fine-tuned methods and our E$^2$STR-ICL. {We fine-tune ABINet \cite{abinet}, SATRN \cite{satrn} and MAERec \cite{Union14M} with the same data preserved in the in-context pool}. As shown in Figure \ref{fig:fine-tuning compare}, E$^2$STR-ICL outperforms MAERec by 1.16$\%$ even if the latter is fine-tuned with in-domain data, which is an exciting result given that E$^2$STR-ICL requires no parameter updating. In a word, our E$^2$STR can be rapidly implemented in a training-free manner in various novel scenarios and even achieves better performance than the fine-tuned SOTA methods.

\begin{table}[t]
\centering
\resizebox{0.48\textwidth}{11.5mm}{
\begin{tabular}{ccccccc}
\hline
                   & \multicolumn{2}{c}{COCO} & \multicolumn{2}{c}{HOST} & \multicolumn{2}{c}{WordArt} \\ \hline
annotation rate & 10$\%$        & 20$\%$       & 10$\%$        & 20$\%$       & 10$\%$         & 20$\%$         \\ \hline
MAERec \cite{Union14M}             & 0           & 0          & 0           & 0          & 0            & 0            \\
w/ fine-tuning     & 0.82        & 1.67       & 1.03        & 1.72       & 1.34         & 2.23         \\ \hline
E$^2$STR-base           & 0           & 0          & 0           & 0          & 0            & 0            \\
E$^2$STR-ICL            & 10.12       & 12.92      & 12. 43      & 13.76      & 26.22        & 32.02        \\ \hline
\end{tabular}
}
\caption{Results on hard case rectification. ``Hard Cases" are test samples misrecognized by both MAERec \cite{Union14M} and our E$^2$STR-base. By providing annotation of a small part of the hard cases, we compare the performance increase in the rest test samples between the fine-tuned MAERec and our E$^2$STR-ICL.}
\label{Table: hard case rectification}
\end{table}

\subsubsection{Results on Hard Case Rectification}
\label{sec: Hard Case Rectification}

We demonstrate the rectification ability of E$^2$STR, which can handle hard cases in STR conveniently and effectively, in a training-free manner. Specifically, we define ``hard cases" as the scene text samples that are wrongly recognized by both E$^2$STR-base and the SOTA method MAERec. A small number of hard cases are then annotated, and we study how the model can benefit from the annotated hard cases and decrease the error rate of the rest hard cases. Shown in Table \ref{Table: hard case rectification}, we perform experiments on COCO Text, HOST, and WordArt. As we can see, by annotating 10$\%$ to 20$\%$ of the hard cases, E$^2$STR-ICL decreases the error rate of the rest hard cases by up to 32$\%$. This improvement is achieved by simply putting the annotated hard cases into the in-context pool, without any hassle of re-training the model. By comparison, by fine-tuning on the annotated hard cases, MAERec only decreases the error rate by up to 2.23$\%$, completely incomparable to our E$^2$STR-ICL. As a result, E$^2$STR-ICL can rapidly learn from hard cases and improve the performance in a training-free manner, while SOTA methods like MAERec can hardly benefit from hard samples even with fine-tuning.

\begin{table}[t]
\centering
\begin{tabular}{cccccc}
\hline
\multicolumn{3}{c}{Training Task} &  & \multicolumn{2}{c}{Word Accuracy} \\ \cline{1-3} \cline{5-6} 
VT        & TS        & SS        &  & Non-Context              & In-Context              \\ \hline
  \checkmark        &           &           &  & 69.69                  &    26.82              \\
  \checkmark        &    \checkmark       &           &  &     69.80      &   75.60               \\
   \checkmark       &           &    \checkmark       &  &          69.66         &       73.09           \\
    \checkmark      &   \checkmark        &    \checkmark       &  &    69.66        &   76.77        \\ \hline
\end{tabular}
\caption{Ablation of our proposed training strategies, where VT, TS, and SS refer to vanilla STR training, the Transform Strategy, and the Split Strategy. The experiment is performed on EIST.}
\label{ablation of training strategy}
\end{table}

\begin{figure}[t]
        \centering
        \includegraphics[width=0.46\textwidth]{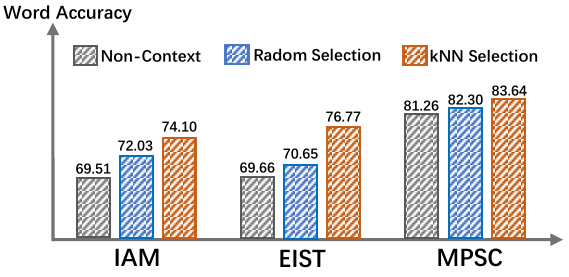}
        \caption{Comparison between different in-context prompt selection strategies. ``Random Selection" refers to randomly selecting two samples as in-context prompts from the in-context pool. {X-axis is the evaluated benchmarks.}}
        \label{knn_ablation}
\end{figure}

\subsection{Ablation Studies}
\label{sec:ablation studies}

\noindent \textbf{Impact of Split-and-Transform Training Strategies.} We perform an experiment to show the effectiveness of our proposed Split Strategy and Transform Strategy. Shown in Table \ref{ablation of training strategy}, the vanilla STR training brings a word accuracy of 69.69$\%$, but the model cannot understand context information, and the performance even severely decreases to 26.82$\%$ when provided with in-context prompts. The application of the Transform Strategy and the Split Strategy in the second training stage does not increase the non-context performance (concerning that the synthesized data is typically weaker than the real-world data used in the vanilla training stage), but the model learns to profit from context, and the performance is improved to 75.60$\%$ and 73.09$\%$ respectively when provided with in-context prompts. Finally, the hybrid of the above two strategies further enhances the ICL ability, and the performance reaches 76.77$\%$.

\noindent \textbf{Impact of Nearest Neighbor In-Context Prompt Selection.} 
In Section \ref{method: Inference} we propose to select samples most similar to the test image in the latent space based on the kNN strategy. Here we demonstrate the effectiveness of this strategy by comparing the performance to Random Selection, {\it i.e.}, randomly selecting in-context prompts from the in-context pool. Shown in Figure \ref{knn_ablation}, on all three evaluated datasets, random selection can improve the performance of non-context prediction by a small margin, but is far from comparing with kNN selection. Specifically, on EIST random selection improves the performance of non-context from 69.66$\%$ to 70.65$\%$, while kNN selection reaches 76.77$\%$ word accuracy under the same condition.

\begin{figure}[t]
        \centering
        \includegraphics[width=0.46\textwidth]{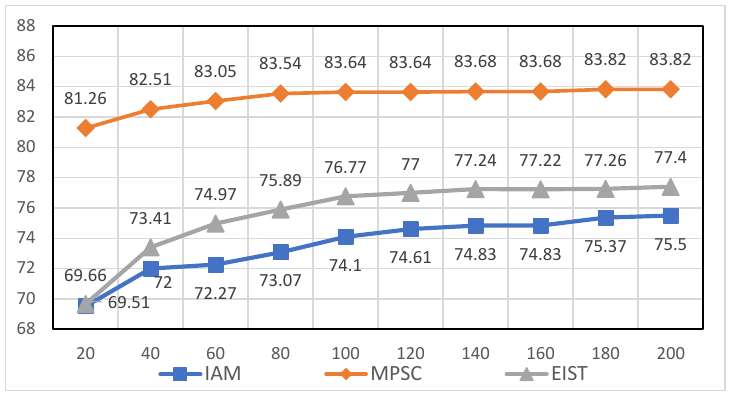}
        \caption{The performance change brought by different sizes of the in-context pool. The X-axis is the size of the in-context pool and the Y-axis is the word accuracy results.}
        \label{pool_size_ablation}
\end{figure}

\noindent \textbf{Impact of In-Context Pool Size.} We next study the impact of varying the size of the in-context pool. Shown in Figure \ref{pool_size_ablation}, we perform experiments on IAM, EIST, and MPSC, by varying the number of samples maintained in the in-context pool. As we can see, in general, the larger in-context pool brings about better performance, and this improvement effect weakens as the pool continually expands. To be specific, on IAM the word accuracy is increased from 69.51$\%$ to 74.10$\%$ (4.59$\%$ improvement) when the pool size is 100, while it only increases the performance from 74.10$\%$ to 75.50$\%$ (1.40$\%$ improvement) when the pool is expanded with another 100 samples. The above fact implies that a small number of samples is adequate to bring about huge performance improvement when deploying E$^2$STR-ICL.

\noindent \textbf{Impact of the Number of In-Context prompts.} We analyze the influence of the number of in-context prompts. Shown in Table \ref{Table: ablation of the number of in-context examples}, the experiment is performed on HOST, ToTal Text, and WordArt. Similar to the in-context pool size, the increase in the number of in-context prompts also generally boosts the performance of E$^2$STR-ICL. However, as we can see, one to two in-context prompts are adequate for improving the performance by a large margin, and the further increase of in-context prompts brings about a limited improvement. This phenomenon is possibly caused by the fact that usually only a few characters are wrongly recognized for a bad case, which can be rectified by the context information from one or two in-context prompts.

\begin{table}[t]
\centering
\resizebox{0.45\textwidth}{8mm}{
\begin{tabular}{ccccccc}
\hline
ICL prompts & 0     & 1     & 2     & 4     & 8     & 16    \\ \hline
HOST                   & 73.30 & 74.34 & 74.75 & 74.83 & 74.83 & 74.92 \\
TT                     & 93.46 & 94.68 & 94.68 & 94.78 & 94.88 & 94.91 \\
WordArt                & 81.47 & 86.04 & 86.17 & 86.17 & 86.27 & 86.35 \\ \hline
\end{tabular}
}
\caption{The performance change brought by the different number of in-context prompts.}
\label{Table: ablation of the number of in-context examples}
\end{table}

\noindent \textbf{Computational Complexity.} We experimentally compare the inference speed of E$^2$STR and MAERec \cite{Union14M}. Shown in Table \ref{table: computational complexity}, the inference speed of E$^2$STR-ICL is on par with MAERec. Compared to E$^2$STR-base, the in-context prompts of E$^2$STR-ICL bring extra consumption, but this leads to a limited inference time increase ({\it i.e.}, from 0.071 to 0.094). It makes sense since we only maintain the visual tokens in the in-context pool and directly feed the visual tokens of the selected prompts to the language model.

\begin{table}[t]
\centering
\resizebox{0.45\textwidth}{4.5mm}{
\begin{tabular}{cccc}
\hline
                   & MAERec \cite{Union14M} & E$^2$STR-base & E$^2$STR-ICL \\ \hline
Inference Time (s) & 0.092  & 0.071    & 0.094   \\ \hline
\end{tabular}
}
\caption{Comparison of the mean inference time of each test sample. All results are reported under the same hardware environment.}
\label{table: computational complexity}
\end{table}

\begin{figure}[t!]
        \centering
        \includegraphics[width=0.48\textwidth]{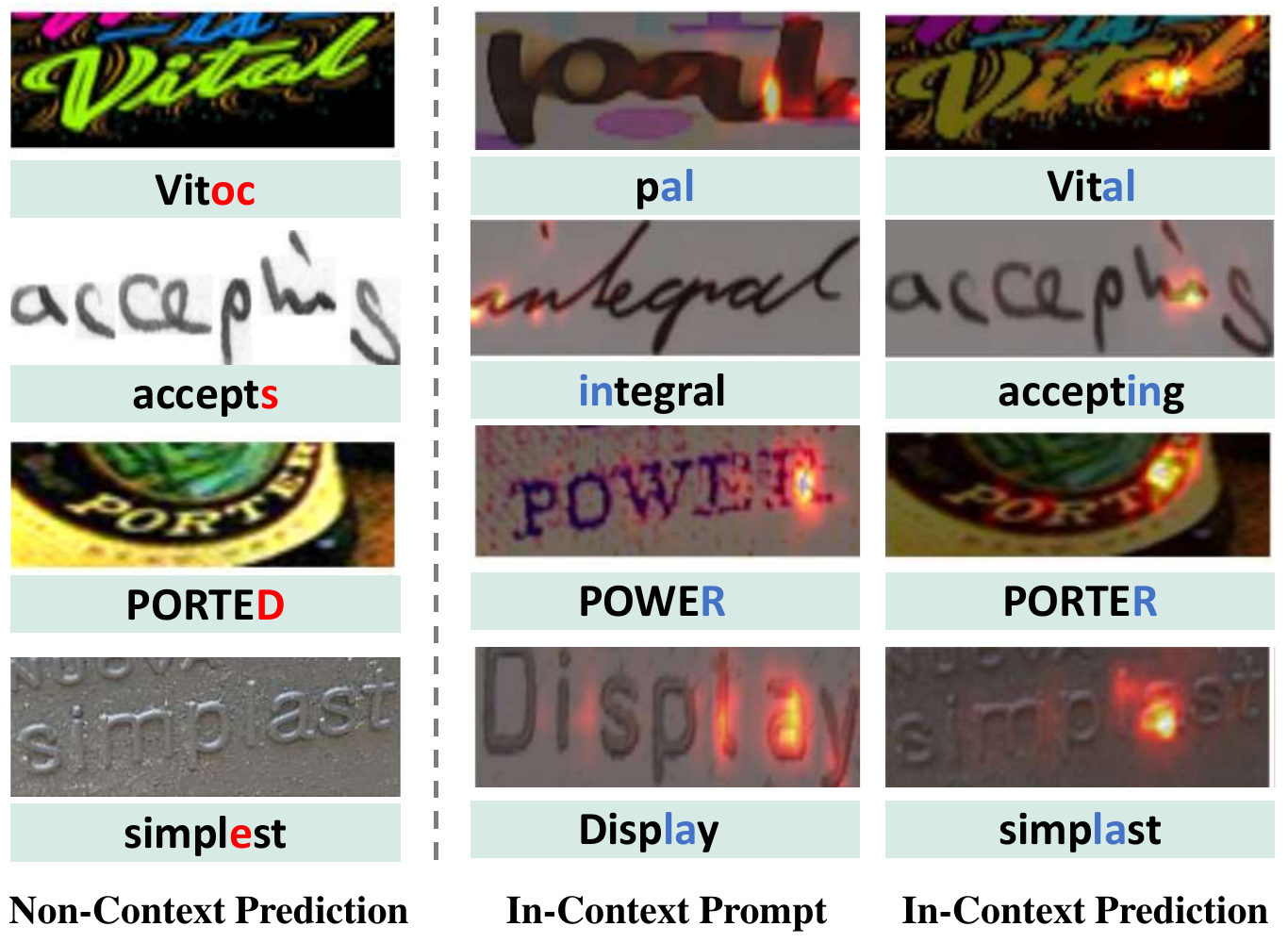}
        \caption{Cross attention visualization between the language tokens and the vision tokens. {Left: Non-context prediction of E$^2$STR. Error characters are marked in {red}. Right: In-context prediction of E$^2$STR-ICL, where only one in-context prompt is selected. We visualize how the language tokens attend to the prompt image and the test image.}}

        \label{fig: heat map}
\end{figure}

\noindent\textbf{Visualization and Further Analysis.} We further study how the test sample learns from context. Shown in Figure \ref{fig: heat map}, we select one context prompt for the test sample, and study the model pays attention to which region of the context image. This is achieved by collecting the attention maps between the language tokens and the image features. As we can see, when the language tokens pay close attention to the misrecognized image region, they also focus on the context image region which has similar patterns. For example, on the last row of Figure \ref{fig: heat map}, E$^2$STR misrecognized the test image as ``simplest" without context. By providing a context prompt ``Display", one language token focuses on the ``la" region of both images, which have similar image patterns. Finally, E$^2$STR rectified the misrecognized ``e" to ``a" with the help of context ground-truth ``la" of the focused region.


\section{Limitations}

There are two limitations in our study. Firstly, there is a very slim chance that E$^2$STR-ICL erroneously rectifies predictions due to misleading prompts (please refer to supplementary materials). Additionally, our model still could not recognize characters that are not included in the lexicon.

\section{Conclusion}

In this paper, we propose E$^2$STR, an ego-evolving scene text recognizer equipped with in-context learning capabilities. Through our proposed in-context training strategy incorporating context-rich scene text sequences, E$^2$STR performs rapid adaptation across diverse scenarios without additional fine-tuning. Extensive experiments demonstrate that E$^2$STR not only achieves SOTA performance on common STR benchmarks but also outperforms even the approaches that have been fine-tuned specifically for cross-domain scenarios.  The model's ability to easily and effectively handle difficult text cases further underscores its potential as a unified text recognizer. Overall, this research represents a significant step toward efficient and highly adaptive text recognition models well-suited for diverse real-world applications.

{
    \small
    \bibliographystyle{ieeenat_fullname}
    \bibliography{main}

\begin{thebibliography}{47}
\providecommand{\natexlab}[1]{#1}
\providecommand{\url}[1]{\texttt{#1}}
\expandafter\ifx\csname urlstyle\endcsname\relax
  \providecommand{\doi}[1]{doi: #1}\else
  \providecommand{\doi}{doi: \begingroup \urlstyle{rm}\Url}\fi

\bibitem[Alayrac et~al.(2022)Alayrac, Donahue, Luc, Miech, Barr, Hasson, Lenc, Mensch, Millican, Reynolds, et~al.]{flamingo}
Jean-Baptiste Alayrac, Jeff Donahue, Pauline Luc, Antoine Miech, Iain Barr, Yana Hasson, Karel Lenc, Arthur Mensch, Katherine Millican, Malcolm Reynolds, et~al.
\newblock Flamingo: a visual language model for few-shot learning.
\newblock \emph{Advances in Neural Information Processing Systems}, 35:\penalty0 23716--23736, 2022.

\bibitem[Baek et~al.(2019)Baek, Kim, Lee, Park, Han, Yun, Oh, and Lee]{internal1}
Jeonghun Baek, Geewook Kim, Junyeop Lee, Sungrae Park, Dongyoon Han, Sangdoo Yun, Seong~Joon Oh, and Hwalsuk Lee.
\newblock What is wrong with scene text recognition model comparisons? dataset and model analysis.
\newblock In \emph{Proceedings of the IEEE/CVF international conference on computer vision}, pages 4715--4723, 2019.

\bibitem[Bautista and Atienza(2022)]{parseq}
Darwin Bautista and Rowel Atienza.
\newblock Scene text recognition with permuted autoregressive sequence models.
\newblock In \emph{European Conference on Computer Vision}, pages 178--196. Springer, 2022.

\bibitem[Belval()]{synthesize}
Belval.
\newblock Generator.
\newblock \url{https://github.com/Belval/TextRecognitionDataGenerator}.

\bibitem[Borisyuk et~al.(2018)Borisyuk, Gordo, and Sivakumar]{ctc1}
Fedor Borisyuk, Albert Gordo, and Viswanath Sivakumar.
\newblock Rosetta: Large scale system for text detection and recognition in images.
\newblock In \emph{Proceedings of the 24th ACM SIGKDD international conference on knowledge discovery \& data mining}, pages 71--79, 2018.

\bibitem[Brown et~al.(2020)Brown, Mann, Ryder, Subbiah, Kaplan, Dhariwal, Neelakantan, Shyam, Sastry, Askell, et~al.]{gpt3}
Tom Brown, Benjamin Mann, Nick Ryder, Melanie Subbiah, Jared~D Kaplan, Prafulla Dhariwal, Arvind Neelakantan, Pranav Shyam, Girish Sastry, Amanda Askell, et~al.
\newblock Language models are few-shot learners.
\newblock \emph{Advances in neural information processing systems}, 33:\penalty0 1877--1901, 2020.

\bibitem[Cheng et~al.(2017)Cheng, Bai, Xu, Zheng, Pu, and Zhou]{internal2}
Zhanzhan Cheng, Fan Bai, Yunlu Xu, Gang Zheng, Shiliang Pu, and Shuigeng Zhou.
\newblock Focusing attention: Towards accurate text recognition in natural images.
\newblock In \emph{Proceedings of the IEEE international conference on computer vision}, pages 5076--5084, 2017.

\bibitem[Ch'ng and Chan(2017)]{total-text}
Chee~Kheng Ch'ng and Chee~Seng Chan.
\newblock Total-text: A comprehensive dataset for scene text detection and recognition.
\newblock In \emph{2017 14th IAPR international conference on document analysis and recognition (ICDAR)}, pages 935--942. IEEE, 2017.

\bibitem[Dosovitskiy et~al.(2020)Dosovitskiy, Beyer, Kolesnikov, Weissenborn, Zhai, Unterthiner, Dehghani, Minderer, Heigold, Gelly, et~al.]{ViT}
Alexey Dosovitskiy, Lucas Beyer, Alexander Kolesnikov, Dirk Weissenborn, Xiaohua Zhai, Thomas Unterthiner, Mostafa Dehghani, Matthias Minderer, Georg Heigold, Sylvain Gelly, et~al.
\newblock An image is worth 16x16 words: Transformers for image recognition at scale.
\newblock \emph{arXiv preprint arXiv:2010.11929}, 2020.

\bibitem[Fang et~al.(2021)Fang, Xie, Wang, Mao, and Zhang]{abinet}
Shancheng Fang, Hongtao Xie, Yuxin Wang, Zhendong Mao, and Yongdong Zhang.
\newblock Read like humans: Autonomous, bidirectional and iterative language modeling for scene text recognition.
\newblock In \emph{Proceedings of the IEEE/CVF Conference on Computer Vision and Pattern Recognition}, pages 7098--7107, 2021.

\bibitem[Graves et~al.(2006)Graves, Fern{\'a}ndez, Gomez, and Schmidhuber]{ctc}
Alex Graves, Santiago Fern{\'a}ndez, Faustino Gomez, and J{\"u}rgen Schmidhuber.
\newblock Connectionist temporal classification: labelling unsegmented sequence data with recurrent neural networks.
\newblock In \emph{Proceedings of the 23rd international conference on Machine learning}, pages 369--376, 2006.

\bibitem[Guan et~al.(2022)Guan, Gu, Lu, Tu, Feng, Wu, and Guan]{mpsc}
Tongkun Guan, Chaochen Gu, Changsheng Lu, Jingzheng Tu, Qi Feng, Kaijie Wu, and Xinping Guan.
\newblock Industrial scene text detection with refined feature-attentive network.
\newblock \emph{IEEE Transactions on Circuits and Systems for Video Technology}, 32\penalty0 (9):\penalty0 6073--6085, 2022.

\bibitem[Gupta and Kembhavi(2023)]{mmicl2}
Tanmay Gupta and Aniruddha Kembhavi.
\newblock Visual programming: Compositional visual reasoning without training.
\newblock In \emph{Proceedings of the IEEE/CVF Conference on Computer Vision and Pattern Recognition}, pages 14953--14962, 2023.

\bibitem[He et~al.(2023{\natexlab{a}})He, Wang, Hu, Liu, Liu, Xu, and Shen]{icl-d3ie}
Jiabang He, Lei Wang, Yi Hu, Ning Liu, Hui Liu, Xing Xu, and Heng~Tao Shen.
\newblock Icl-d3ie: In-context learning with diverse demonstrations updating for document information extraction.
\newblock \emph{arXiv preprint arXiv:2303.05063}, 2023{\natexlab{a}}.

\bibitem[He et~al.(2023{\natexlab{b}})He, Wang, Hu, Liu, Liu, Xu, and Shen]{mmicl4}
Jiabang He, Lei Wang, Yi Hu, Ning Liu, Hui Liu, Xing Xu, and Heng~Tao Shen.
\newblock Icl-d3ie: In-context learning with diverse demonstrations updating for document information extraction.
\newblock \emph{arXiv preprint arXiv:2303.05063}, 2023{\natexlab{b}}.

\bibitem[He et~al.(2022)He, Chen, Xie, Li, Doll{\'a}r, and Girshick]{MAE}
Kaiming He, Xinlei Chen, Saining Xie, Yanghao Li, Piotr Doll{\'a}r, and Ross Girshick.
\newblock Masked autoencoders are scalable vision learners.
\newblock In \emph{Proceedings of the IEEE/CVF conference on computer vision and pattern recognition}, pages 16000--16009, 2022.

\bibitem[Jiang et~al.(2023)Jiang, Wang, Peng, Liu, and Jin]{Union14M}
Qing Jiang, Jiapeng Wang, Dezhi Peng, Chongyu Liu, and Lianwen Jin.
\newblock Revisiting scene text recognition: A data perspective.
\newblock In \emph{Proceedings of the IEEE/CVF International Conference on Computer Vision}, pages 20543--20554, 2023.

\bibitem[Karatzas et~al.(2013)Karatzas, Shafait, Uchida, Iwamura, i~Bigorda, Mestre, Mas, Mota, Almazan, and De~Las~Heras]{ic13}
Dimosthenis Karatzas, Faisal Shafait, Seiichi Uchida, Masakazu Iwamura, Lluis~Gomez i Bigorda, Sergi~Robles Mestre, Joan Mas, David~Fernandez Mota, Jon~Almazan Almazan, and Lluis~Pere De~Las~Heras.
\newblock Icdar 2013 robust reading competition.
\newblock In \emph{2013 12th international conference on document analysis and recognition}, pages 1484--1493. IEEE, 2013.

\bibitem[Karatzas et~al.(2015)Karatzas, Gomez-Bigorda, Nicolaou, Ghosh, Bagdanov, Iwamura, Matas, Neumann, Chandrasekhar, Lu, et~al.]{ic15}
Dimosthenis Karatzas, Lluis Gomez-Bigorda, Anguelos Nicolaou, Suman Ghosh, Andrew Bagdanov, Masakazu Iwamura, Jiri Matas, Lukas Neumann, Vijay~Ramaseshan Chandrasekhar, Shijian Lu, et~al.
\newblock Icdar 2015 competition on robust reading.
\newblock In \emph{2015 13th international conference on document analysis and recognition (ICDAR)}, pages 1156--1160. IEEE, 2015.

\bibitem[Lee et~al.(2020)Lee, Park, Baek, Oh, Kim, and Lee]{satrn}
Junyeop Lee, Sungrae Park, Jeonghun Baek, Seong~Joon Oh, Seonghyeon Kim, and Hwalsuk Lee.
\newblock On recognizing texts of arbitrary shapes with 2d self-attention.
\newblock In \emph{Proceedings of the IEEE/CVF Conference on Computer Vision and Pattern Recognition Workshops}, pages 546--547, 2020.

\bibitem[Li et~al.(2023)Li, Zhang, Chen, Wang, Yang, and Liu]{otter}
Bo Li, Yuanhan Zhang, Liangyu Chen, Jinghao Wang, Jingkang Yang, and Ziwei Liu.
\newblock Otter: A multi-modal model with in-context instruction tuning.
\newblock \emph{arXiv preprint arXiv:2305.03726}, 2023.

\bibitem[Li et~al.(2019)Li, Wang, Shen, and Zhang]{sar}
Hui Li, Peng Wang, Chunhua Shen, and Guyu Zhang.
\newblock Show, attend and read: A simple and strong baseline for irregular text recognition.
\newblock In \emph{Proceedings of the AAAI conference on artificial intelligence}, pages 8610--8617, 2019.

\bibitem[Liao et~al.(2019)Liao, Zhang, Wan, Xie, Liang, Lyu, Yao, and Bai]{segmentation1}
Minghui Liao, Jian Zhang, Zhaoyi Wan, Fengming Xie, Jiajun Liang, Pengyuan Lyu, Cong Yao, and Xiang Bai.
\newblock Scene text recognition from two-dimensional perspective.
\newblock In \emph{Proceedings of the AAAI conference on artificial intelligence}, pages 8714--8721, 2019.

\bibitem[Liu et~al.(2016)Liu, Chen, Wong, Su, and Han]{ctc2}
Wei Liu, Chaofeng Chen, Kwan-Yee~K Wong, Zhizhong Su, and Junyu Han.
\newblock Star-net: a spatial attention residue network for scene text recognition.
\newblock In \emph{BMVC}, page~7, 2016.

\bibitem[Liu et~al.(2019)Liu, Jin, Zhang, Luo, and Zhang]{ctw}
Yuliang Liu, Lianwen Jin, Shuaitao Zhang, Canjie Luo, and Sheng Zhang.
\newblock Curved scene text detection via transverse and longitudinal sequence connection.
\newblock \emph{Pattern Recognition}, 90:\penalty0 337--345, 2019.

\bibitem[Ma et~al.(2021)Ma, Wang, Huang, Huang, Goulermas, and Huang]{visual-understanding-1}
Mengkai Ma, Qiu-Feng Wang, Shan Huang, Shen Huang, Yannis Goulermas, and Kaizhu Huang.
\newblock Residual attention-based multi-scale script identification in scene text images.
\newblock \emph{Neurocomputing}, 421:\penalty0 222--233, 2021.

\bibitem[Marti and Bunke(2002)]{iam}
U-V Marti and Horst Bunke.
\newblock The iam-database: an english sentence database for offline handwriting recognition.
\newblock \emph{International Journal on Document Analysis and Recognition}, 5:\penalty0 39--46, 2002.

\bibitem[Mei et~al.(2020)Mei, Hu, Yang, Zheng, and Hu]{industrial-print-1}
Qiang Mei, Qinyou Hu, Chun Yang, Hailin Zheng, and Zhisheng Hu.
\newblock Port recommendation system for alternative container port destinations using a novel neural language-based algorithm.
\newblock \emph{IEEE Access}, 8:\penalty0 199970--199979, 2020.

\bibitem[Mishra et~al.(2012)Mishra, Alahari, and Jawahar]{IIIT5k}
Anand Mishra, Karteek Alahari, and CV Jawahar.
\newblock Top-down and bottom-up cues for scene text recognition.
\newblock In \emph{2012 IEEE conference on computer vision and pattern recognition}, pages 2687--2694. IEEE, 2012.

\bibitem[OUALI et~al.(2022)OUALI, HALIMA, and Ali]{augmented-1}
Imene OUALI, Mohamed~BEN HALIMA, and WALI Ali.
\newblock Augmented reality for scene text recognition, visualization and reading to assist visually impaired people.
\newblock \emph{Procedia Computer Science}, 207:\penalty0 158--167, 2022.

\bibitem[Phan et~al.(2013)Phan, Shivakumara, Tian, and Tan]{svtp}
Trung~Quy Phan, Palaiahnakote Shivakumara, Shangxuan Tian, and Chew~Lim Tan.
\newblock Recognizing text with perspective distortion in natural scenes.
\newblock In \emph{Proceedings of the IEEE international conference on computer vision}, pages 569--576, 2013.

\bibitem[Risnumawan et~al.(2014)Risnumawan, Shivakumara, Chan, and Tan]{cute80}
Anhar Risnumawan, Palaiahankote Shivakumara, Chee~Seng Chan, and Chew~Lim Tan.
\newblock A robust arbitrary text detection system for natural scene images.
\newblock \emph{Expert Systems with Applications}, 41\penalty0 (18):\penalty0 8027--8048, 2014.

\bibitem[Saudagar and Mohammad(2018)]{augmented-2}
Abdul Khader~Jilani Saudagar and HabeebVulla Mohammad.
\newblock Augmented reality mobile application for arabic text extraction, recognition and translation.
\newblock \emph{Journal of Statistics and Management Systems}, 21\penalty0 (4):\penalty0 617--629, 2018.

\bibitem[Shao et~al.(2023)Shao, Yu, Wang, and Yu]{mmicl3}
Zhenwei Shao, Zhou Yu, Meng Wang, and Jun Yu.
\newblock Prompting large language models with answer heuristics for knowledge-based visual question answering.
\newblock In \emph{Proceedings of the IEEE/CVF Conference on Computer Vision and Pattern Recognition}, pages 14974--14983, 2023.

\bibitem[Sheng et~al.(2019)Sheng, Chen, and Xu]{nrtr}
Fenfen Sheng, Zhineng Chen, and Bo Xu.
\newblock Nrtr: A no-recurrence sequence-to-sequence model for scene text recognition.
\newblock In \emph{2019 International conference on document analysis and recognition (ICDAR)}, pages 781--786. IEEE, 2019.

\bibitem[Shi et~al.(2018)Shi, Yang, Wang, Lyu, Yao, and Bai]{aster}
Baoguang Shi, Mingkun Yang, Xinggang Wang, Pengyuan Lyu, Cong Yao, and Xiang Bai.
\newblock Aster: An attentional scene text recognizer with flexible rectification.
\newblock \emph{IEEE transactions on pattern analysis and machine intelligence}, 41\penalty0 (9):\penalty0 2035--2048, 2018.

\bibitem[Shi et~al.(2014)Shi, Wang, Xiao, Gao, and Hu]{svt}
Cunzhao Shi, Chunheng Wang, Baihua Xiao, Song Gao, and Jinlong Hu.
\newblock End-to-end scene text recognition using tree-structured models.
\newblock \emph{Pattern Recognition}, 47\penalty0 (9):\penalty0 2853--2866, 2014.

\bibitem[Tsimpoukelli et~al.(2021)Tsimpoukelli, Menick, Cabi, Eslami, Vinyals, and Hill]{frozen}
Maria Tsimpoukelli, Jacob~L Menick, Serkan Cabi, SM Eslami, Oriol Vinyals, and Felix Hill.
\newblock Multimodal few-shot learning with frozen language models.
\newblock \emph{Advances in Neural Information Processing Systems}, 34:\penalty0 200--212, 2021.

\bibitem[Veit et~al.(2016)Veit, Matera, Neumann, Matas, and Belongie]{coco-text}
Andreas Veit, Tomas Matera, Lukas Neumann, Jiri Matas, and Serge Belongie.
\newblock Coco-text: Dataset and benchmark for text detection and recognition in natural images.
\newblock \emph{arXiv preprint arXiv:1601.07140}, 2016.

\bibitem[Wan et~al.(2020)Wan, He, Chen, Bai, and Yao]{segmantation2}
Zhaoyi Wan, Minghang He, Haoran Chen, Xiang Bai, and Cong Yao.
\newblock Textscanner: Reading characters in order for robust scene text recognition.
\newblock In \emph{Proceedings of the AAAI conference on artificial intelligence}, pages 12120--12127, 2020.

\bibitem[Wang et~al.(2021)Wang, Xie, Fang, Wang, Zhu, and Zhang]{ost}
Yuxin Wang, Hongtao Xie, Shancheng Fang, Jing Wang, Shenggao Zhu, and Yongdong Zhang.
\newblock From two to one: A new scene text recognizer with visual language modeling network.
\newblock In \emph{Proceedings of the IEEE/CVF International Conference on Computer Vision}, pages 14194--14203, 2021.

\bibitem[Xie et~al.(2022)Xie, Fu, Zhang, Wang, and Bai]{wordart}
Xudong Xie, Ling Fu, Zhifei Zhang, Zhaowen Wang, and Xiang Bai.
\newblock Toward understanding wordart: Corner-guided transformer for scene text recognition.
\newblock In \emph{European Conference on Computer Vision}, pages 303--321. Springer, 2022.

\bibitem[Yang et~al.(2022)Yang, Gan, Wang, Hu, Lu, Liu, and Wang]{mmicl7}
Zhengyuan Yang, Zhe Gan, Jianfeng Wang, Xiaowei Hu, Yumao Lu, Zicheng Liu, and Lijuan Wang.
\newblock An empirical study of gpt-3 for few-shot knowledge-based vqa.
\newblock In \emph{Proceedings of the AAAI Conference on Artificial Intelligence}, pages 3081--3089, 2022.

\bibitem[Yu et~al.(2020)Yu, Li, Zhang, Liu, Han, Liu, and Ding]{SRN}
Deli Yu, Xuan Li, Chengquan Zhang, Tao Liu, Junyu Han, Jingtuo Liu, and Errui Ding.
\newblock Towards accurate scene text recognition with semantic reasoning networks.
\newblock In \emph{Proceedings of the IEEE/CVF conference on computer vision and pattern recognition}, pages 12113--12122, 2020.

\bibitem[Zhang et~al.(2021)Zhang, Tao, Du, Ding, Wang, Liu, and Wang]{driving-1}
Chongsheng Zhang, Yuefeng Tao, Kai Du, Weiping Ding, Bin Wang, Ji Liu, and Wei Wang.
\newblock Character-level street view text spotting based on deep multisegmentation network for smarter autonomous driving.
\newblock \emph{IEEE Transactions on Artificial Intelligence}, 3\penalty0 (2):\penalty0 297--308, 2021.

\bibitem[Zhang et~al.(2022)Zhang, Roller, Goyal, Artetxe, Chen, Chen, Dewan, Diab, Li, Lin, et~al.]{OPT}
Susan Zhang, Stephen Roller, Naman Goyal, Mikel Artetxe, Moya Chen, Shuohui Chen, Christopher Dewan, Mona Diab, Xian Li, Xi~Victoria Lin, et~al.
\newblock Opt: Open pre-trained transformer language models.
\newblock \emph{arXiv preprint arXiv:2205.01068}, 2022.

\bibitem[Zhao et~al.(2023)Zhao, Cai, Si, Ma, An, Chen, Liu, Wang, Han, and Chang]{mmicl1}
Haozhe Zhao, Zefan Cai, Shuzheng Si, Xiaojian Ma, Kaikai An, Liang Chen, Zixuan Liu, Sheng Wang, Wenjuan Han, and Baobao Chang.
\newblock Mmicl: Empowering vision-language model with multi-modal in-context learning.
\newblock \emph{arXiv preprint arXiv:2309.07915}, 2023.

\end{thebibliography}
}

\clearpage
\setcounter{page}{1}
\maketitlesupplementary

\section{Model Architecture}

\begin{figure}[h]
        \centering
        \includegraphics[width=0.48\textwidth]{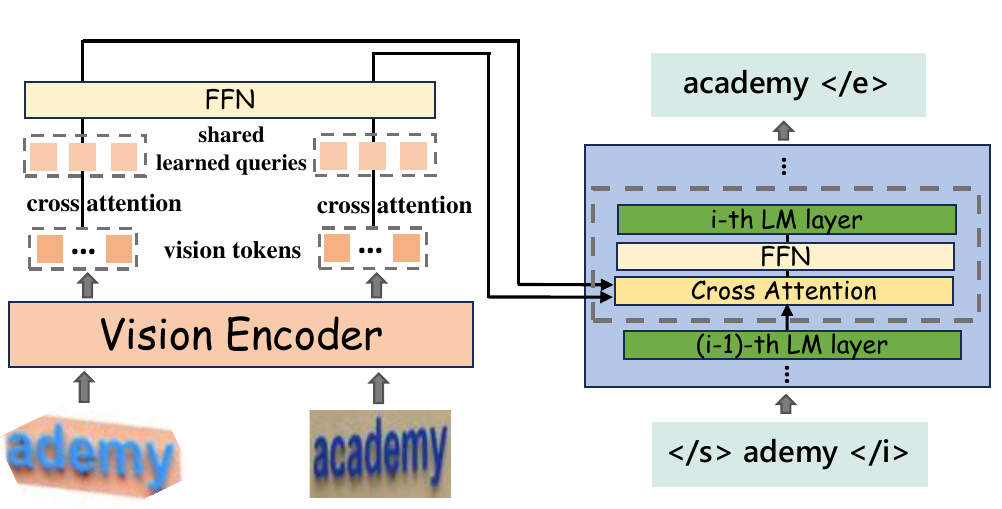}
        \caption{Detailed Model Architecture of E$^2$STR.}

        \label{fig: detailed architecture}
\end{figure}

Figure \ref{fig: detailed architecture} presents the detailed model architecture of E$^2$STR. We follow the paradigm established by Flamingo \cite{flamingo}, where we perform cross attention between the vision outputs and the language outputs in each language model layer. The language outputs serve as queries and the vision outputs serve as keys and values. The detailed configures of the vision encoder and the language decoder are summarized in Table \ref{Table: model details}. For fair comparison, we provide MAERec \cite{Union14M} with the same language decoder with E$^2$STR-ICL (We name this modification as MAERec$^{\dag}$). The comparison between MAERec$^{\dag}$ and E$^2$STR is shown in Table \ref{Table: maerec same decoder}.

\begin{table}[h]
\centering
\resizebox{0.5\textwidth}{10.5mm}{
\begin{tabular}{ccccccc}
\hline
                                                           & \begin{tabular}[c]{@{}c@{}}Input\\ Size\end{tabular} & \begin{tabular}[c]{@{}c@{}}Patch\\ Size\end{tabular} & Embedding & Depth & Heads & Parameters \\ \hline
\begin{tabular}[c]{@{}c@{}}Vision\\ Encoder\end{tabular}   & 32x128                                                & 4x4                                                  & 768       & 12    & 12    & 85M        \\ \hline
\begin{tabular}[c]{@{}c@{}}Language\\ Decoder\end{tabular} & -                                                    & -                                                    & 768       & 12    & 12    & 125M       \\ \hline
\end{tabular}
}
\caption{Model details of E$^2$STR.}
\label{Table: model details}
\end{table}

\begin{table}[h]
\centering
\begin{tabular}{cccc}
\hline
                & MPSC  & EIST  & IAM   \\ \hline
MAERec          & 81.81 & 70.33 & 70.27 \\
MAERec$^{\dag}$ & 82.00 & 70.77 & 70.51 \\
E$^2$STR-ICL    & 83.64 & 76.77 & 74.10 \\ \hline
\end{tabular}
\caption{Word Accuracy performance comparison between MAERec \cite{Union14M} and E$^2$STR-ICL. MAERec$^{\dag}$ refers to MAERec using the same vision encoder and the same language decoder with E$^2$STR-ICL.}
\label{Table: maerec same decoder}
\end{table}

\section{Model Scalability}

Table \ref{time change by prompt} presents the inference time change brought by the different number of in-context prompts. It is easy to find that the number of in-context prompts in E$^2$STR is scalable. For example, the inference time of E$^2$STR-ICL (where we select two prompts) is 0.094s. But When expanding the number of in-context prompts by 7 times ({\it i.e.}, 16 prompts), the inference time is only increased by 1.08 times ({\it i.e.}, 0.196s).

\begin{table}[h]
\centering
\resizebox{0.5\textwidth}{5.5mm}{
\begin{tabular}{ccccccc}
\hline
Prompts            & 0     & 1     & 2     & 4     & 8     & 16    \\ \hline
Inference Time (s) & 0.071 & 0.085 & 0.094 & 0.113 & 0.140 & 0.196 \\ \hline
\end{tabular}
}
\caption{Inference time change brought by the different number of in-context prompts.}
\label{time change by prompt}
\end{table}

Table \ref{time change by pool} presents the inference time change brought by different sizes of the in-context pool. As we can see, when expanding the pool size by 4 times ({\it i.e.}, from 100 to 500), the inference time is only increased by 0.07 times ({\it i.e.}, from 0.094 to 0.101). As a result, our E$^2$STR-ICL is highly scalable in terms of both in-context pool size and the number of in-context prompts.

\begin{table}[h]
\centering
\resizebox{0.5\textwidth}{5.5mm}{
\begin{tabular}{cccccc}
\hline
Pool Size          & 100   & 200   & 300   & 400   & 500   \\ \hline
Inference Time (s) & 0.094 & 0.096 & 0.097 & 0.099 & 0.101 \\ \hline
\end{tabular}
}
\caption{Inference time change brought by different sizes of the in-context pool.}
\label{time change by pool}
\end{table}

\begin{table}[h]
\centering
\begin{tabular}{ccccc}
\hline
     & \multicolumn{4}{c}{Prompt Domain}                              \\ \cline{2-5} 
     & Non-context & MPSC           & EIST           & IAM            \\ \hline
MPSC & 81.26       & \textbf{83.64} & 83.00          & 82.96          \\
EIST & 69.66       & 70.30          & \textbf{76.77} & 70.00          \\
IAM  & 69.51       & 72.17          & 71.70          & \textbf{74.10} \\ \hline
\end{tabular}
\caption{Performance change brought by the domain variation of the in-context pool. \textbf{Bold} values denote the best performance in a row.}
\label{performance change by domain pool}
\end{table}

\section{Model Stability}

\begin{figure}[h]
        \centering
        \includegraphics[width=0.48\textwidth]{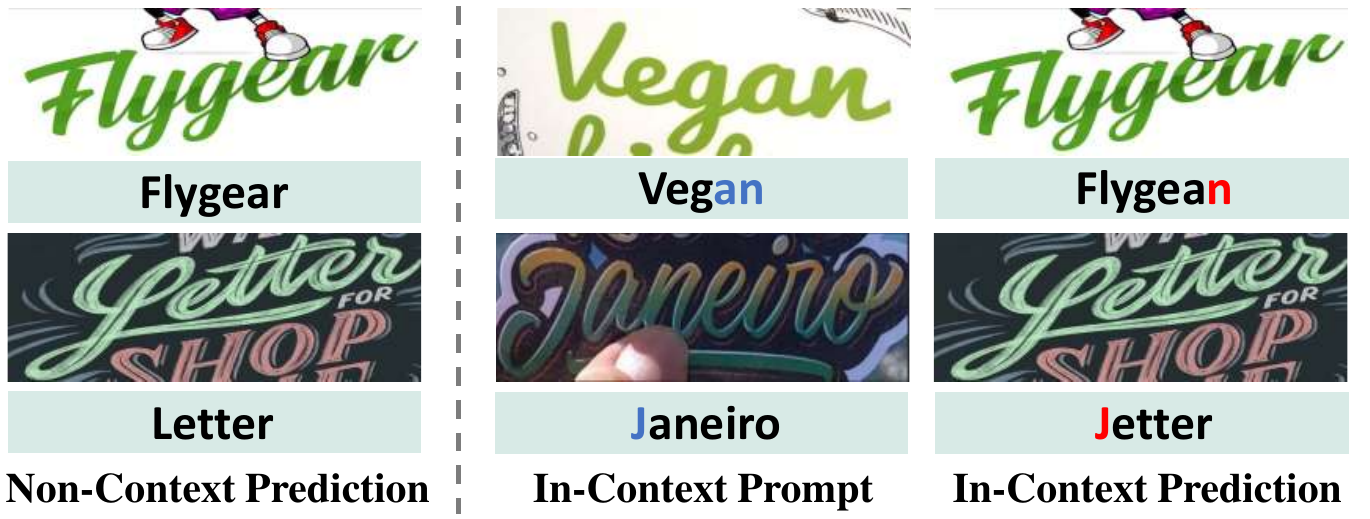}
        \caption{Examples of erroneous rectification brought by misleading prompts.}

        \label{fig: misleading}
\end{figure}

Table \ref{performance change by domain pool} presents how the performance change when varying the domains of the in-context pool. As we can see, our E$^2$STR-ICL is stable to the change of the context prompts. On all three benchmarks, out-of-domain in-context pools still improve the performance, though the improvement is lower than in-domain in-context pools. Nevertheless, there still exists a very slim chance that E$^2$STR-ICL erroneously rectifies predictions due to misleading prompts. Shown in Figure \ref{fig: misleading}, when certain areas of the prompt image is highly similar to the test image but the ground-truth is different, E$^2$STR-ICL may erroneously rectifies the prediction.

\section{Visualization}

\begin{figure}[h]
        \centering
        \includegraphics[width=0.48\textwidth]{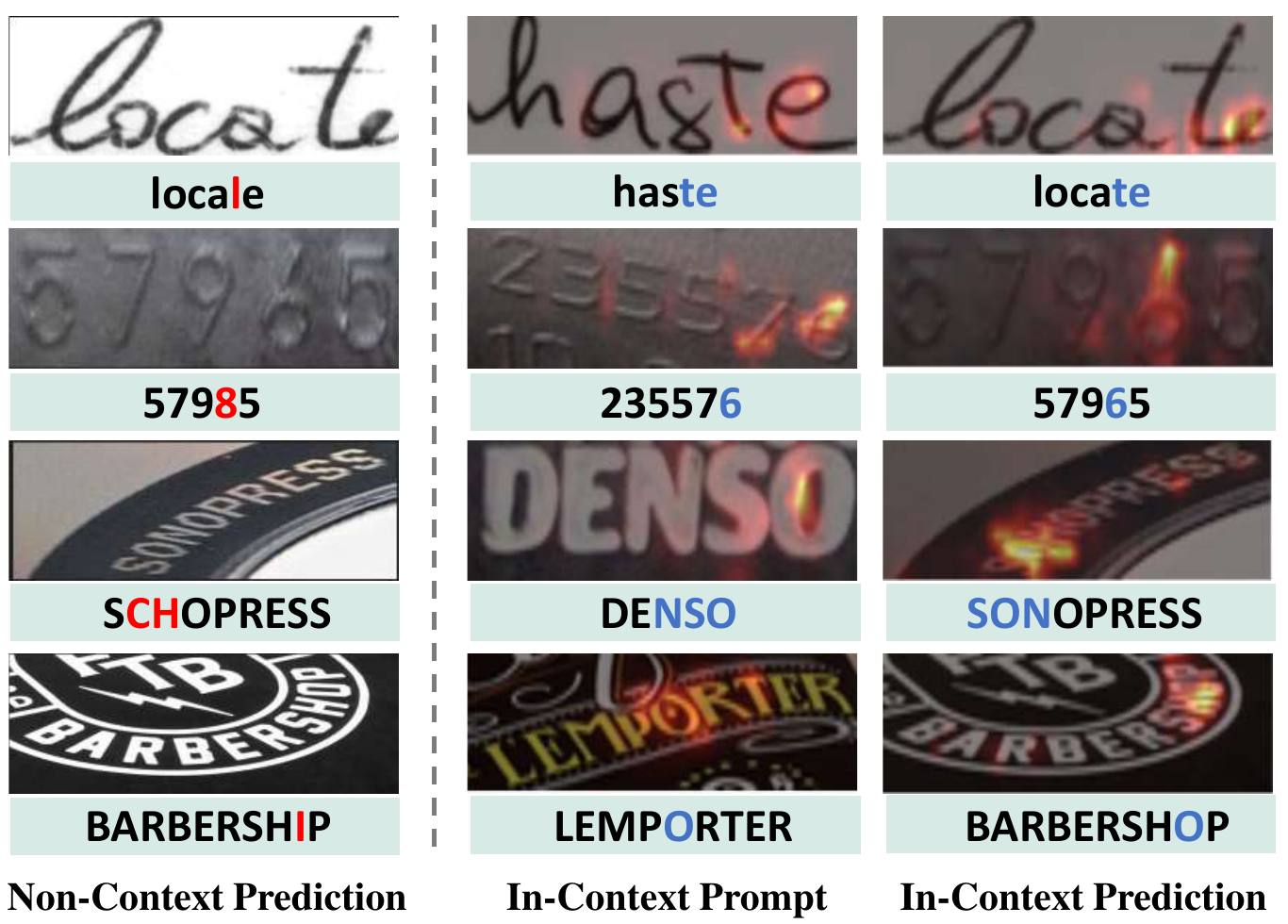}
        \caption{More examples of the cross attention visualization.}

        \label{fig: more heat map}
\end{figure}

We provide more examples of the cross attention visualization in Figure \ref{fig: more heat map}.

\begin{table}[t]
\centering

\begin{tabular}{ccccccc}
\hline
                     & \begin{tabular}[c]{@{}c@{}}Training\\ GPU Hours\end{tabular} &      & MPSC  & EIST  & IAM   & AVG   \\ \hline
\multirow{2}{*}{kNN} & \multirow{2}{*}{\textbf{415.6}}                                       & base & 81.22 & 69.78 & 69.62 & 73.54 \\
                     &                                                              & ICL  & 82.06 & 70.95 & 71.00 & \textbf{74.67} \\ \hline
\multirow{2}{*}{ST}  & \multirow{2}{*}{\textbf{131.2}}                                       & base & 81.26 & 69.66 & 69.51 & 73.48 \\
                     &                                                              & ICL  & 83.64 & 76.77 & 74.10 & \textbf{78.17} \\ \hline
\end{tabular}

\vspace{-0.3cm}
\caption{Comparisons between kNN and our ST-strategy during in-context training.}
\vspace{-0.2cm}
\label{knn-compare}
\end{table}

\end{document}